\documentclass[conference]{IEEEtran}
\IEEEoverridecommandlockouts

\usepackage{cite}
\usepackage{url}
\usepackage{amsmath,amssymb,amsfonts}
\usepackage{graphicx}
\usepackage{caption}
\usepackage{subcaption}
\usepackage{textcomp}
\usepackage{xcolor}
\usepackage{booktabs}
\usepackage{multirow}
\usepackage{algorithm}
\usepackage[noend]{algpseudocode}

\def\BibTeX{{\rm B\kern-.05em{\sc i\kern-.025em b}\kern-.08em
    T\kern-.1667em\lower.7ex\hbox{E}\kern-.125emX}}
\begin{document}

\title{Anomaly Detection in Time Series Data Using Reinforcement Learning, Variational Autoencoder, and Active Learning

}

\author{\IEEEauthorblockN{1\textsuperscript{st} Bahareh Golchin}
\IEEEauthorblockA{\textit{dept. Computer Science} \\
\textit{Portland State University}\\
Portland, Oregon \\
bgolchin@pdx.edu}
\and
\IEEEauthorblockN{2\textsuperscript{nd} Banafsheh Rekabdar}
\IEEEauthorblockA{\textit{dept. Computer Science} \\
\textit{Portland State University}\\
Portland, Oregon \\
rekabdar@pdx.edu}

}
% Configuration for caption style
\captionsetup{
    font=footnotesize, % Set the font size to footnote size (which is roughly 8pt)
    labelsep=space, % Separate label and text with a space
    justification=centering % Center the caption text
}

\maketitle 
\begin{abstract}

A novel approach to detecting anomalies in time series data is presented in this paper. This approach is pivotal in domains such as data centers, sensor networks, and finance. Traditional methods often struggle with manual parameter tuning and cannot adapt to new anomaly types. Our method overcomes these limitations by integrating Deep Reinforcement Learning (DRL) with a Variational Autoencoder (VAE) and Active Learning. By incorporating a Long Short-Term Memory (LSTM) network, our approach models sequential data and its dependencies effectively, allowing for the detection of new anomaly classes with minimal labeled data. Our innovative DRL-VAE and Active Learning combination significantly improves existing methods, as shown by our evaluations on real-world datasets, enhancing anomaly detection techniques and advancing time series analysis.

\end{abstract}

\begin{IEEEkeywords}
Anomaly detection, Deep reinforcement learning, Variational autoencoder, Active learning, Long short-term memory, Generative AI
\end{IEEEkeywords}

\section{Introduction}
Detecting anomalies in time series plays a key role in different areas, such as data centers, sensor networks, cyber-physical systems, and finance \cite{b1,b2,b3,b4}. Manual tuning of parameters and features and specific data properties is required in most of the existing methods in the literature. Furthermore, due to large amounts of data, manually identifying anomalies in time series data: 1) takes a lot of time and labor, and 2) is likely to be influenced by human errors. Therefore, an automated system is needed to detect anomalies in extensive time series data \cite{b5}.

Detecting anomalies presents two main challenges: First, because anomalies are rare, it is difficult to train models effectively to spot them. Second, the fact that real-world data often changes over time adds another layer of complexity. In response to the scarcity of labeled data, many anomaly detection algorithms have been suggested typically unsupervised, which do not require labeled data. These algorithms are usually based on specific assumptions about anomaly patterns observed in the data. However, these assumptions may not always align with real-world situations, leading to high false-positive rates. This mismatch arises from varying user interests and definitions of anomalies \cite{b6, b7}.

Supervised methods are highly effective when sufficient labeled data is available but face difficulties in scenarios with limited or no labels. 
These methods assume that the underlying distribution stays the same, even when labels are available. If the distribution changes, they need to be retrained\cite{b8}.

Next, the approach utilized to detect anomalies in time series data is semi-supervised learning. This method is not suitable for situations where there are many different types of anomalies because it only works with a small number of known anomalies, and it can miss new anomalies in data that has not been labeled. Thus, this approach cannot detect new types of anomalies \cite{b9}.

To tackle the issues mentioned above, namely the problem of weakly-supervised anomaly detection in time series data, we utilize Deep Reinforcement Learning (DRL). The exploration versus exploitation dilemma is crucial in Reinforcement Learning (RL) \cite{b10}. Taking actions that are best based on what is already known is called exploitation in the literature. However, exploration is trying new things to find better actions. The agent must balance between acting optimally with current knowledge and seeking more knowledge.

The DRL used in our study is developed to efficiently utilize a small portion of labeled anomalous data ($D_{la}$). This approach extensively explores a large pool of unlabeled data ($D_u$), which detects new classes of anomalies not covered by the labeled data. This exploration is enhanced by the integration of a Variational Autoencoder (VAE), which powers the DRL framework. 

Furthermore, recognizing the high cost and scarcity of fully labeled data in real-world scenarios, active learning has been integrated into our system which enables our RL agent to 1) explore the environment and accumulate experience effectively, and 2) make informed queries based on this experience during its exploration.

Long Short-Term Memory (LSTM) network is the core of our DRL agent. This 1) simulates sequential time series data, and 2) extracts the long-term dependencies between activities \cite{b11}.

To summarize, our key contributions include the following:
\begin{itemize}
    \item To the best of our knowledge, the combination of DRL-VAE and an Active Learning approach (i.e., RLVAL) to detect anomalies in time series data proposed in this paper is one of the first studies in the literature. 
    \item We investigate the use of LSTMs to enhance the robustness of time series data modeling and integrate these networks into our DRL framework.
    \item We evaluate our approach on two time series datasets (i.e., Yahoo and KPI). Our results demonstrate that RLVAL surpasses previous state-of-the-art methods.
   
\end{itemize}
In what follows, we lay out the structure of the rest of this paper. In Section II, we review the studies in the literature which is related to our work. Moreover, the background in anomaly detection is explored. Next, our proposed framework is detailed in Section III. Section IV discusses the implementation, including a comprehensive examination of the datasets used. Finally, we conclude our study in Section V.

\section{Related Work}

Recently, detecting anomalies has been the focus of numerous methodologies. Broadly, we can classify these methods as follows. 1) Statistical-based methods, and 2) machine learning-based approaches.
\subsection{Statistical-based Methods}
Statistical-based models involve building a model from given datasets, and then, using mathematical tests to decide if the unseen data fits the proposed model. Kernel function-based methods directly learn normal behavior from the training data \cite{b12,b13}. Parametric statistical models, such as Gaussian, regression, and logistic regression models, assume the underlying distribution of normal data follows an existing distribution \cite{b14}. However, these methods assume that the normal behavior fits an existing distribution which is not fair in practice.
\subsection{Machine Learning-based Methods}
Machine learning-based approaches use labeled training data to differentiate between normal and abnormal data, achieving this through either classification or clustering techniques. Common algorithms include Bayesian networks \cite{b15}, support vector machines \cite{b16}, rule-based systems \cite{b17}, and neural networks \cite{b18}. Clustering algorithms, such as k-means are also used to detect anomalies \cite{b19}.

Next, for complex time series data, specialized techniques and algorithms have been developed. Examples include 1) Skyline \cite{b20}, which detects anomalies in real-time, and 2) Twitter's package, which detects anomalies where seasonality and trends are present \cite{b21}.

Contextual anomaly detection, exemplified by ContextOSE \cite{b22}, focuses on capturing local information rather than global patterns. Hierarchical temporal memory (HTM), seen in projects like Numenta and Numenta TM, stores and recalls temporal and spatial patterns \cite{b23}.

Along with the advancement of deep learning, to detect anomalies through time series data, Recurrent Neural Networks (RNN) or LSTM models have been developed. These models learn from normal training data to predict future values and detect anomalies based on prediction errors. Variants based on autoencoders have also been explored \cite{b24}.

Recently, RL has attracted attention to detect anomalies in time series data. The reason is that it has a generic framework, and it can learn from itself. For instance, convolutional autoencoders within an RL framework to detect anomalies are proposed by Bourdonnaye et al. \cite{b25}. Similarly, a value-based DRL approach using the Deep Q-Network (DQN) algorithm is proposed by Huang et al. \cite{b26}.

To expand the literature, a new combination of the DQN algorithm with autoencoders and active learning is proposed. This method creates a more robust model for identifying anomalies in time series data.
\section{Background}

Before introducing our proposed method, we provide an overview of key concepts such as RL, DQNs, VAE, and Active Learning to understand our approach better. 
\subsection{Reinforcement Learning in Anomaly Detection}\label{AA}

In this study, we define detecting anomalies challenge as a Markov Decision Process (MDP). This MDP is represented as the tuple $<S, A, P_a, R_a, \gamma>$. Here, $S$ represents the set of possible states within the environment, while $A$ includes the actions available to the RL agent. The transition probability $R_a(s, s\;')$ indicates the likelihood of moving from state $s$ to state $ s\;'$ under action $a$. Moreover, $R_a(s, s\;')$ is the immediate reward received after transitioning from $s$ to $ s\;'$ through action $a$. $\gamma$ represents the discount factor, which ranges from 0 to 1, and it diminishes the value of future rewards.

We define $V_\pi(s)$ as the value function. $V_\pi(s)$ represents the expected return from state $s$, which is calculated as follows.
\begin{equation}
   V_\pi(s)=E\left[\sum_{t=0}^{\infty} \gamma^t R_t \mid s_0 = s\right]
\end{equation}
forecasting the total reward accrued from starting at state $s$ under policy $\pi$. The agent aims to maximize the cumulative future reward by learning a policy $\pi: S \to A$.

Model-based or model-free methods are addressed in MDP. The former involves constructing a detailed model of the environment, requiring the agent to understand and interact with it effectively, with dynamic programming as a prominent example. This approach breaks down complex problems into simpler, manageable subproblems. In contrast, model-free methods do not need a comprehensive understanding of the environment. They rather focus on exploring and predicting subsequent states to determine optimal actions. Since they are more widely applicable, our discussion will focus only on model-free methods.

In the model-free approach, we differentiate between two main strategies: value-based and policy-based algorithms. These methods are central to our analysis and further exploration.

\subsection{Deep Q-Networks and Q-Learning}
One of the value-based RL algorithms is $Q$-learning. The agent in this algorithm learns the action-value function, $Q(s, a)$. The value of taking a specific action at a given state is predicted using this function. The target value for updates is defined as:
\begin{equation}
   \text{target} = R_{s,a,s'} + \gamma \max_{a'} Q_k(s', a')
\end{equation}
The following formula shows how the $Q$-function is updated.
\begin{equation}
   Q_{k+1}(s, a) \leftarrow (1 - \alpha)Q_k(s, a) + \alpha \text{target}
\end{equation}

However, traditional $Q$-learning can become unstable or even diverge. This could take place particularly when the action-value function is approximated utilizing nonlinear functions such as neural networks \cite{b27}.

To overcome these challenges, $DeepMind$ developed a method called DQN, which combines RL with deep neural networks to handle more complex problems effectively. DQN improves the action-value function approximation by introducing two key ideas: 1) experience replay and 2) a target network \cite{b28}. Experience replay stores a history of state transitions, each of which is recorded as a tuple  $<s, a, r, s\;'>$. 

DQN allows the agent to train on a diverse set of experiences, reducing correlations between consecutive samples and increasing the efficiency of the learning process. The target network helps stabilize learning by providing a fixed baseline for the target values for a period, facilitating smoother updates and helping the main network to converge.

\subsection{Variational Autoencoder}
VAEs model the transformation between original feature spaces and simpler latent Gaussian distributions. In this process, 1) the feature space is converted into Gaussian distributions using encoders, and 2) the feature space from these distributions is reconstructed using decoders. Both components are implemented using neural networks. Maximizing the marginal likelihood \( p(x; \theta) \) is the main goal of a VAE. In this context, 1) \( x \) denotes a feature vector, 2) all the parameters of the decoder \( p(x \mid z; \theta) \) are captured in \( \theta \), and \( z \) represents the latent space.\\

Inadvertently, we cannot trace the marginal likelihood. Therefore, we use the Evidence Lower Bound (ELBO) as an approximation. The ELBO is formulated as the following \cite{b5}.
\begin{equation}
   L(\theta, \phi; x) = \langle \log p(x \mid z; \theta) \rangle_{q(z \mid x; \phi)} - KL[q(z \mid x; \phi) \| p(z)]
\end{equation}

where \( L(\theta, \phi; x) \leq \log p(x; \theta) \). In this context, \( q(z \mid x; \phi) \) denotes the encoder, parameterized by \(\phi\). Next, \( KL[\cdot \| \cdot] \) is defined as the Kullback-Leibler divergence. Finally, \(\langle \cdot \rangle_{p(\cdot)}\) represents the expectation over the distribution \(p(\cdot)\).

%To improve training, the expectation in the ELBO is approximated by using a finite number of $L$ samples of $z$, expressed as:
%\begin{equation}
%  L(\theta, \phi; x) \approx \frac{1}{L} \sum_{l=1}^{L} \log p(x \mid z^{(l)}; \theta) - KL[q(z \mid x; \phi) \| p(z)]
%\end{equation},
%\begin{equation}
%  z^{(l)} \sim q(z \mid x; \phi), \quad l \in \{1, 2, \ldots, L\}
%end{equation}

%where $\sim$ signifies a sample from the specified distribution. Typically, a standard Gaussian prior is used, which is represented by $p(z) = N(z; 0, 1)$. Further, a linear combination constant $C$ may be added to enhance performance.

\subsection{Active Learning in Machine Learning Systems}

Active learning utilizes users to enhance learning efficiency. This machine learning technique primarily requests specific data from the user that it deems beneficial for learning. Consider a scenario where a labeled training set is represented by
$L=(X,Y)$, and a pool of unlabeled instances is defined by $U = (x_1, x_2,...,x_n)$. As unlabeled data is typically less costly than labeled data, the pool $U$ can be substantial in size.

The essence of active learning lies in its ability to selectively query unlabeled instances from $U$ using a query function $Q$, and request manual labeling by human experts. This selective process targets samples that are deemed to be most informative to the current model $C$, thus maximizing the learning impact from the newly labeled instances. The newly labeled dataset $L_{new} = (X_{new}, Y_{new})$ is then incorporated into the training set $P$ for subsequent training iterations. Active learning aims to refine the classifier model $C$ efficiently, utilizing a minimal number of queries.

There are several querying strategies within active learning, each tailored to different data acquisition needs:
\begin{itemize}
    \item Random Selection: Samples are randomly chosen from $U$ and added to $L$.
    \item Least Confidence: This strategy selects samples for which the model $C$ has the lowest confidence in its predictions. The sample $x_{lc}$ is chosen such that:
    \begin{equation}
    x_{lc} = \arg \max (1 - P_C(\hat{y} \mid x))
    \end{equation}
    where $\hat{y}$ is the label with the highest predicted probability by model $C$, indicating the model's uncertainty.

    \item Margin Sampling: Samples are chosen based on the smallest difference in model confidence between the two most probable class predictions:
    \begin{equation}
    x_m = \arg \min (P_C(\hat{y}_1 \mid x) - P_C(\hat{y}_2 \mid x))
    \end{equation}
     where $\hat{y_1}$ and $\hat{y_2}$ are the first and second most likely class labels predicted by model $C$.
     \item Entropy Sampling: This approach selects samples that have the highest entropy in their prediction distributions, indicative of greater informational value:
     \begin{equation}
    x_E = \arg \max_x \left(- \sum P_C(y_i \mid x) \log P_C(y_i \mid x)\right)
    \end{equation}
   
\end{itemize}

Each of these strategies aims to optimize the learning process by focusing on the acquisition of the most informative data, which improves the training phase.

\section{PROPOSED METHOD}

In this section, a detailed description of each component of our approach RLVAL is provided. Our proposed method integrates DRL with a VAE and incorporates Active Learning into our framework. Fig. 1 illustrates our entire proposed method.
\begin{figure*}[t]
    \centering
    \includegraphics[width=0.8\textwidth]{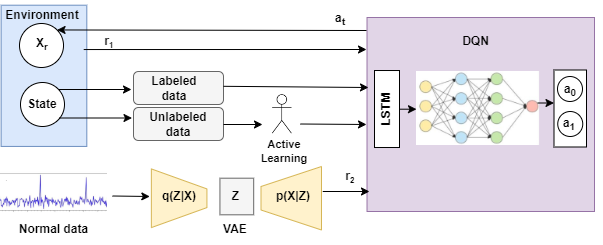} 
    \caption{Overview of RLVAL system}
    \label{fig:widefigure}
\end{figure*}
\subsection{Anomaly Detection with Variational Autoencoders}
VAEs have been effectively applied to anomaly detection as an unsupervised learning method. This demonstrates the VAEs' ability to learn representations from feature vectors efficiently. The primary mechanism for detecting anomalies using a VAE is connected to analyzing how large the reconstruction loss is. This proposed VAE is trained exclusively on normal data. Anomalous samples, because of their nature, are excluded from this training set.

When evaluating unlabeled samples, the VAE attempts to reconstruct each sample. Samples that are normal are typically reconstructed with minimal loss, indicating their conformity to the learned normal patterns. In contrast, anomalous samples tend to cause higher reconstruction losses due to their deviation from these patterns. By setting a specific threshold for reconstruction loss, this metric can be used as an anomaly score. Samples that yield a reconstruction loss exceeding this threshold are then classified as anomalies. This method parallels traditional outlier detection techniques, where deviations from established norms are flagged as potential anomalies.
\begin{figure}[b]
\centerline{\includegraphics{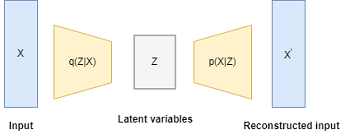}}
\caption{The VAE framework}
\label{fig}
\end{figure}
Fig. 2 illustrates the way VAE is used in our framework.

\subsection{Deep Reinforcement Learning Framework}
We adopted an RL approach, namely DQN, in our framework due to its capacity for generalization and its ability for incremental self-learning. Our DQN framework aims to balance exploiting a small labeled dataset ($D_{la}$) with exploring new anomaly classes in a larger unlabeled dataset ($D_u$). The labeled data can potentially enhance detection accuracy. Moreover, the search for predefined anomalies is eliminated.

Our exploration strategy applies a VAE, which uses a large volume of unlabeled data to provide a supervisory signal to facilitate the unsupervised detection of anomalies. Unlike traditional methods that might rely solely on autoencoders for unsupervised learning, our use of VAEs allows for more powerful anomaly detection by comparing the reconstruction of normal and abnormal sequences, thereby generating an anomaly score.

The core of our framework is the integration of VAE into a weakly supervised learning setup, where it signals deviations from normalcy without direct training on specific anomalies. 
Our primary objective is for the agent to be led through interactions with the environment that is built on training data. This will allow the agent to identify and explore potential anomalies beyond known examples.

Our environment is designed to support both the exploitation of known anomalies and the exploration of new unlabeled data through a mixed reward function. This function can balance exploitation and exploration by using the combination of these labeled anomalies and suspicious unlabeled data. During training, this helps the agent improve its understanding of abnormalities and make informed decisions about new data instances.

The RL setup incorporates three main components: 1) the agent, 2) the environment, and 3) the reward system. The agent begins at a specific time step. It takes actions based on its current state and policy. It, then, receives feedback as a reward. The agent takes action so that the cumulative discounted rewards are maximized. The optimal policy is described mathematically as the following:

\begin{equation}
   \pi^* = \arg\max_{\pi} E_{\pi} \left[ \sum_{t=0}^{\infty} \gamma^t r_t \right]
\end{equation}

where 
$\gamma \text{ (0} \leq \gamma \leq \text{1)}$ represents the discount factor, and it weighs the importance of future rewards.

To evaluate the effectiveness of each action, we utilize the state-action-value function (i.e., Q-value function) to obtain optimal policy.
This state-action-value function is formulated as follows.
\begin{equation}
   Q_{\pi}(s, a) = E_{\pi} \left[ \sum_{T=t}^{\infty} \gamma^{T-t} r_T \mid s_t = s, a_t = a \right]
\end{equation}

This function is crucial to update the policy using the Bellman equation:
\begin{equation}
   Q_{\pi}(s, a) \leftarrow Q_{\pi}(s, a) + \alpha \left( r + \gamma \max_{a'} Q(s', a') - Q(s, a) \right)
\end{equation}

where $s'$ is the new state, $a'$ is the new action, and $\alpha$ is the learning rate.

A key strategy in our framework is the use of experience replay, which stores past state transitions and enables efficient learning by breaking correlations between consecutive samples. This method not only increases data efficiency but also helps in stabilizing updates by reducing variance.

The exploration versus exploitation dilemma has always been a critical aspect of RL. To resolve this dilemma exploiting actions on existing knowledge and exploring new strategies must be balanced. If this balance takes place, future rewards are maximized. Therefore, our proposed framework is designed to navigate this balance, which optimizes the learning process as well as enhances the detection capabilities of the DRL agent.

The reward function $X_r$ in our framework assigns an extrinsic reward $r_1$ to the agent based on the actions taken and the state observed and it is defined as follows:\\ 

$r_1 = $
$\begin{cases} 
1 & \text{if } a = a_1 \text{ and } s \in D_{la}, \\
0 & \text{if } a = a_0 \text{ and } s \in D_{u}, \\
-1 & \text{otherwise}.
\end{cases}$

This setup encourages the agent to maximize the use of $D_{la}$ while remaining neutral towards $D_u$ with the VAE driving how the unlabeled data should be explored.

On top of the extrinsic reward $r_1$, an intrinsic reward $r_2$ from the VAE is received by the agent. This helps the agent to explore potential anomalies in $D_u$ on its own. The VAE evaluates the abnormality of each time window by the size of the reconstruction error, where it utilizes this measure as the intrinsic reward. In our framework, the loss function of the VAE is crucial, and it includes a term for reconstruction loss (i.e., expected negative log-likelihood) for each sample. This function is calculated as follows.
\begin{equation}
L_i(\theta, \phi) = -E_{z \sim q_{\theta}(z | x_i)} [\log p_{\phi}(x_i | z)] + KL(q_{\theta}(z | x_i) \| p(z))
\end{equation}

Following the literature, we define the reconstruction error as the difference between the original input, which is represented by $x$, and its reconstructed counterpart, which is represented by $x'$. The reconstructed error is calculated as the following.
\begin{equation}
\|x - x'\|^2
\end{equation}

We, then, normalize the reconstruction error to improve detecting sensitivity when the probability of anomaly increases with respect to the intrinsic reward $r_2$.  

The overall reward for each time window is defined as follows:
\begin{equation}
r=r_1+r_2
\end{equation}
combines these extrinsic and intrinsic rewards. Therefore, $r$ results in balancing the exploitation of labeled data ($D_{la}$) and exploration of the unlabeled set ($D_u$).

\subsection{Active Learning}

In real-world scenarios, obtaining fully labeled data can be costly. To address this, we incorporated an active learning module into our framework. This addition enhances our RL agent's ability to both navigate through and learn from the environment as well as to generate queries based on its accumulated experiences during these explorations.

We opted for margin sampling as our active learning technique. With smaller margins resulting from this method, we can classify the samples as anomaly or non-anomaly. The set of unlabeled instances, $S_{unlabeled}$, is processed using an active learning module within each episode, which is then received from the DRL framework. During each epoch, the RL agent, at any given state $s$ can choose between two actions: $a_0$ for non-anomaly and $a_1$ for anomaly. These choices are quantified by their respective $Q$-values, calculated as follows:
\begin{equation}
   (q_0, q_1) = W \cdot s + b
\end{equation}

These $Q$-values estimate the potential rewards the RL agent might receive for taking specific actions.

The minimum margin is defined as:
\begin{equation}
   \text{min\_margin} = \min |q_0 - q_1|.
\end{equation}

We compute the margin using Equation 15 and arrange these margins in descending order. A subset of $D_{al}$ instances with the smallest $min\_margin$ values are then reviewed by a human expert. We operate under the assumption that the expert's evaluations are error-free, and thus, the labels are deemed accurate. Subsequently, these newly labeled samples are incorporated into the propagation process. They are, then, added to the sample pool for the next training iteration. In this procedure, humans are involved. Then, we can consider these human-involved labels in our overall count of utilized labels.

Fig. 3 illustrates two sample time windows from the A1Benchmark dataset used as inputs for our model, where the RL model performs actions and receives rewards.

\begin{figure*}[t] 
    \centering
    \begin{subfigure}[b]{0.45\textwidth}
        \centering
        \includegraphics[width=\textwidth]{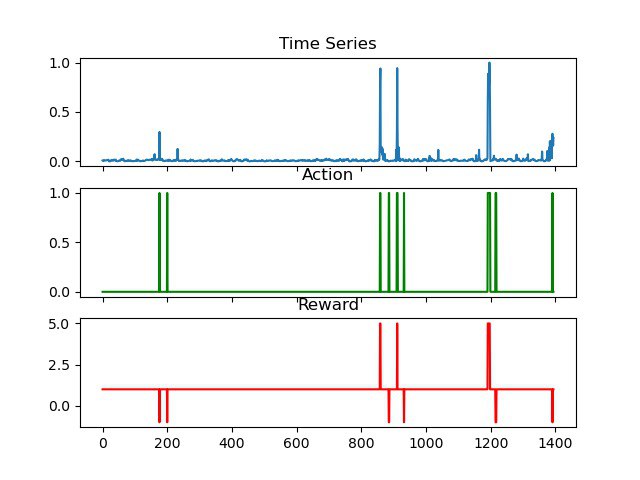} 
        \caption{First time window}
        \label{fig:sub1}
    \end{subfigure}
    \hfill % Horizontal fill to add space between the subfigures if desired
    \begin{subfigure}[b]{0.45\textwidth} 
        \centering
        \includegraphics[width=\textwidth]{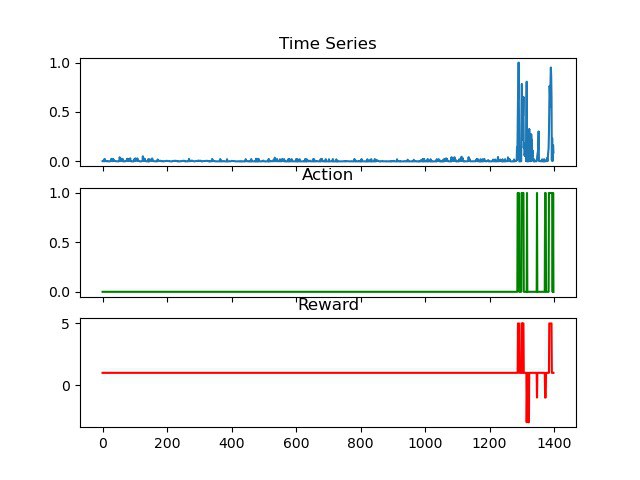} 
        \caption{Second time window}
        \label{fig:sub2}
    \end{subfigure}
    \caption{Sample time windows from the A1Benchmark dataset demonstrating how the RL model processes inputs to perform actions and receive rewards.}
    \label{fig:combined}
\end{figure*}

Algorithm 1 details our proposed model, namely RLVAL.

\begin{algorithm}
\caption{DRL with VAE and Active Learning}
\begin{algorithmic}[1]
\Require Environment; set of states \( S \); replay memory \( D \) of DQN; two same structured neural networks \( Eval_N \) with \( Q \) and \( Target_N \) with \( \hat{Q} \); parameter update ratio \( r \); greedy factor \( \epsilon_t \); discount factor \( \gamma \); learning rate \( \alpha \); pre-trained VAE model \( V \);
\Ensure Action set \( A \);
\State Initial value function \( Q \);
\For{\( E \) in episodes}
    \State Receive labeled instance set \( S \); send unlabeled instance set \( S_{unlabeled} \) to Active Learning and Label Propagation module;
    \For{\( i \) in epochs}
        \State Take a state \( s \) from \( S \);
        \State Generate normal data \( \tilde{s} \) using VAE: \( \tilde{s} = V(s) \);
        \State Compute q\_value = \( Q(\tilde{s}, a) \);
        \State Compute \( Prob_{action} \) based on q\_value and \( \epsilon \);
        \State Action \( a \) = random choice with \( Prob_{action} \);
        \State Observe extrinsic reward \( r_1 \) from the environment,
        \Statex \hspace{2.5em} next state \( s' \);
        \State Compute intrinsic reward \( r_2 \) from VAE;
        \State Total reward \( r = r_1 + r_2 \);
        \State Save transition \( \langle s, a, r, s' \rangle \) in \( D \);
        \State Randomly select a mini-batch \( \langle s_j, a_j, r_j, s'_j \rangle \) 
        \Statex \hspace{2.8em}from \( D \);
        \State Target = \( r_j + \gamma \max \hat{Q}(s_j, a_j) \);
        \State Perform gradient descent: \( (q\_value - target)^2 \);
        \If{\( i \% r == 0 \)}
            \State Copy parameters to \( Target_N \);
        \EndIf
        \State Train VAE during RL training using \( s \) and \( s' \) to \Statex \hspace{2.7em} update \( V \);
    \EndFor
\EndFor
\end{algorithmic}
\end{algorithm}

\section{Experiments}
\subsection{Datasets}
For our study, we conducted evaluations using two well-known datasets: the Yahoo Benchmark and KPI, which are frequently utilized in the analysis of time series anomalies. Table I provides a summary, with more in-depth descriptions.
 \paragraph {Yahoo Benchmark}
   This dataset is designed by Yahoo's webscope project to detect anomalies in time series. It consists of actual traffic data from Yahoo services as well as synthetic datasets. we use the real-time A1Benchmark in our study. This part of the dataset includes Yahoo membership login data and the synthetic dataset. The A1Benchmark consists of 67 time series, each labeled at every timestamp and containing between 1400 and 1600 data points. 
\paragraph {KPI} 
    The KPI dataset originates from the AIOps (Artificial Intelligence for IT Operations) competition and aggregates data from various internet companies, including Tencent, eBay, and Sogou. It encompasses over 3 million data points, each accompanied by timestamps and labels, making it a substantial resource for anomaly detection research.

\begin{table}[t]
\caption{ \\ OVERVIEW OF DATASETS}
    \centering
    \begin{tabular}{|c||c|c|}
        \hline
        
        dataset & total points & anomalies\\
         
        \hline
        Yahoo A1 & 94866 & 1669\\
        \hline
        KPI &3004066 &79554 \\
        
        \hline
    \end{tabular}
    \label{tab1}
\end{table}

\subsection{Metrics}
To compare the performance of our model with other methods in the literature, we use the following three standard metrics.
These metrics are 1) Precision, which evaluates the correctness of the predicted anomalies, 2) Recall, which assesses the coverage of actual anomalies detected by our system, and 3) F1-score, which harmonizes Precision and Recall. 

The formula for the F1-score metric is as follows:

    \begin{equation}
   \text{F1-score} = 2 \times \frac{\text{Precision} \times \text{Recall}}{\text{Precision} + \text{Recall}}
\end{equation}

\subsection{ Results and Discussion}
In this section, our proposed model (i.e., RLVAL) is compared with unsupervised and semi-supervised time series anomaly detection methods.\\
\begin{itemize}
    \item SPOT: Designed for detecting anomalies in streaming time series with one variable. This method automatically determines threshold levels. It requires a preliminary data fraction for initial setup or calibration. In our experiments, we maintained a data split ratio of 80:20, similar to our dataset partitions \cite{b29}.
    \item SR-CNN: This method enhances training by injecting synthetic anomalies into additional training data. It utilizes the Spectral Residual (SR) technique for detecting anomalies, leveraging a custom-trained neural network\cite{b30}.
    \item Autoencoder: This approach uses RNNs with three hidden layers to assess data. The neural network architecture focuses on identifying deviations in data records, providing a quantitative measure of anomalies \cite{b31}.
    \item RLAD: This method is a combination of DRL and active learning to detect anomalies through time series data \cite{b8}.
\end{itemize}

In this paper, We evaluated the Yahoo dataset with three different active queries, namely 1, 5, and 10  (i.e., 1\%, 5\%, and 10\% of data), for each episode. In the KPI dataset, we used 5 and 10 queries (i.e., 0.05\% and 0.1\% of data) in each episode.

We extensively evaluated various anomaly detection techniques on the A1Benchmark from the Yahoo dataset in Table II, underscoring the effectiveness of our proposed method against both unsupervised models such as SPOT, SR-CNN, and Autoencoder, and semi-supervised models like RLAD. Our approach consistently outperformed these models, achieving F1-scores of 0.834 with 1\% labeled data and 0.921 with 10\% labeled data. In contrast, the best-performing RLAD model only reached an F1-score of 0.797 with 10\% labeled data.

Our evaluation on the KPI dataset in Table III showed that RLVAL significantly outperformed traditional unsupervised and semi-supervised techniques. It achieved F1-scores of 0.825 with 0.05\% labeled data and 0.908 with 0.1\% labeled data, demonstrating excellent use of minimal labeled data. In contrast, unsupervised methods like SPOT, SR-CNN, and Autoencoder had F1-scores below 0.170. Even the semi-supervised RLAD model, though better, only reached an F1-score of 0.778 with 0.1\% labeled data.

Our method showed excellent precision and recall, reduced false positives, and increased true anomaly detection. This performance highlights its potential to set new benchmark datasets for anomaly detection, particularly when labeled data is scarce.
\section{Conclusions}
This paper introduced a robust time series anomaly detection method using DRL, VAE, and Active Learning (RLVAL) for a more adaptive, automated system. Our approach uses VAE to enhance the reward received from DRL. Simultaneously, we use active learning to label large amounts of unlabeled data to improve anomaly detection. This combination of DRL, VAE, and active learning enables the model to learn from minimal labeled data and adapt to new data patterns. Our evaluation on Yahoo and KPI datasets shows that our framework outperforms traditional methods. For future work, we suggest combining our model with Large Language Models (LLMs).

\begin{table}[t]
%\centering
\caption{\\COMPARISON OF RLVAL WITH OTHER APPROACHES ON THE YAHOO DATASET}
\label{tab:performance_metrics}
\begin{center}
\begin{tabular}{|l|c|c|c|} 
\hline
\multicolumn{4}{|c|}{A1Benchmark Dataset} \\
\hline
\cline{2-4}
Method & F1-score & Precision & Recall \\
\hline
\multicolumn{4}{|c|}{Unsupervised Learning} \\
\hline
SPOT & \textbf{0.446} & 0.513 & 0.394 \\
SR-CNN & 0.264 & 0.174 & 0.540 \\
Autoencoder & 0.026 & 0.013 & 0.774 \\
\hline
\multicolumn{4}{|c|}{Semi-supervised Learning} \\
\hline
RLAD (1\%) & 0.708 & 0.652 & 0.781 \\
RLAD (5\%) & 0.752 & 0.710 & 0.800 \\
RLAD (10\%) & \textbf{0.797} & 0.733 & 0.922 \\
\hline
%\multicolumn{4}{|c|}{} \\
\hline
RLVAL (1\%) (our approach) & 0.834 & 0.819 & 0.850 \\
RLVAL (5\%) (our approach) & 0.872 & 0.846 & 0.900 \\
RLVAL (10\%) (our approach) & \textbf{0.921}($\uparrow$0.124) & 0.894 & 0.950 \\
\hline
\end{tabular}
\end{center}
\end{table}

\begin{table}[t]
\centering
\caption{\\COMPARISON OF RLVAL WITH OTHER APPROACHES ON THE KPI DATASET}
\label{tab:performance_metrics}
\begin{tabular}{|l|c|c|c|} 
\hline
\multicolumn{4}{|c|}{KPI Dataset} \\
\hline
\cline{2-4}
Method & F1-score & Precision & Recall \\
\hline
\multicolumn{4}{|c|}{Unsupervised Learning} \\
\hline
SPOT & 0.033 & 0.545 & 0.017 \\
SR-CNN & 0.166 & 0.195 & 0.145 \\
Autoencoder & \textbf{0.170} & 0.094 & 0.858 \\
\hline
\multicolumn{4}{|c|}{Semi-supervised Learning} \\
\hline
RLAD (0.05\%) & 0.709 & 0.681 & 0.897 \\
RLAD (0.1\%) & \textbf{0.778} & 0.827 & 0.879 \\
\hline
%\multicolumn{4}{|c|}{} \\
\hline
RLVAL (0.05\%) (our approach) & 0.825 & 0.852 & 0.80 \\
RLVAL (0.1\%) (our approach) & \textbf{0.908} ($\uparrow$0.13) & 0.870 & 0.95 \\
\hline
\end{tabular}
\end{table}

\end{document}